\documentclass[runningheads]{llncs}
\usepackage{graphicx}
\usepackage[misc]{ifsym}
\usepackage{url}
\usepackage{orcidlink}

\usepackage[utf8]{inputenc}
\usepackage[english]{babel}
\usepackage{bbding}
\usepackage{hyperref}
\usepackage{rotating}
\usepackage{verbatim}
\usepackage{amssymb}
\usepackage{amsmath}
\usepackage{xcolor}
\usepackage[most]{tcolorbox}
\usepackage{mathtools, nccmath}
\definecolor{commentgreen}{RGB}{2,112,10}
\definecolor{eminence}{RGB}{108,48,130}
\definecolor{weborange}{RGB}{255,165,0}
\definecolor{frenchplum}{RGB}{129,20,83}
\definecolor{sparqlblue}{RGB}{84,153,199}

\definecolor{framegray}{RGB}{241, 245, 249}
\definecolor{lightgreen}{RGB}{31, 122, 85}
\definecolor{darkgreen}{RGB}{3, 79, 59}
\definecolor{lightblue}{RGB}{57,115,178}
\definecolor{backcolour}{RGB}{240, 253, 250}
\definecolor{bluekeywords}{RGB}{0,0,147}
\definecolor{variable}{RGB}{117,20,124}

\usepackage{listings}

\lstdefinelanguage{SPARQL}{
  morekeywords=[1]{
    SELECT, CONSTRUCT, ASK, DESCRIBE,
    WHERE, FROM, NAMED, OPTIONAL, FILTER,
    UNION, GRAPH, PREFIX, BASE, LIMIT, OFFSET,
    ORDER, BY, ASC, DESC, DISTINCT, REDUCED,
    INSERT, DELETE, WITH, USING, LOAD, CLEAR,
    CREATE, DROP, COPY, MOVE, ADD, DATA, ALL,
    GROUP, HAVING, VALUES, BIND, MINUS, SERVICE,
    SILENT, INTO, AS, NOT, IN, EXISTS
  },
  sensitive=true,
  morecomment=[l]{\#},
  morestring=[b]",
  morestring=[b]',
}

\hypersetup{
    colorlinks,
    citecolor=blue,
    filecolor=black,
    linkcolor=blue,
    urlcolor=blue,
    bookmarksopen=true
}

\usepackage{multicol}

\usepackage{amsthm}
\theoremstyle{definition}
\usepackage[flushleft]{threeparttable}
\usepackage[font=small,labelfont=bf]{caption}
\usepackage[normalem]{ulem}
\begin{document}
\title{CEON: Circular Economy Ontology Network}

%

\author{Huanyu Li \inst{1}  \textsuperscript{(\Letter)}
\orcidlink{0000-0003-1881-3969} 
\and Els de Vleeschauwer \inst{2}
\orcidlink{0000-0002-8630-3947}
\and Robin Keskisärkkä \inst{1}
\orcidlink{0000-0003-0644-4051}
\and \\ Mikael Lindecrantz \inst{3}
\orcidlink{0000-0002-5525-6439}
\and Mina Abd Nikooie Pour \inst{1,4}
\orcidlink{0000-0002-4936-0889}
\and Ying Li \inst{1,4}
\orcidlink{0000-0002-1367-9679}
\and \\ Ben De Meester \inst{2}
\orcidlink{0000-0003-0248-0987}
\and Patrick Lambrix \inst{1,4} 
\orcidlink{0000-0002-9084-0470}
\and Eva Blomqvist \inst{1} \textsuperscript{(\Letter)}  
\orcidlink{0000-0003-0036-6662}
}
\institute{
Department of Computer and Information Science, \\
Link{\"o}ping University, Link{\"o}ping, Sweden 
\\ \email{\{huanyu.li, eva.blomqvist\}@liu.se}
\and IDLab, Department of Electronics and Information Systems, \\ Ghent University – imec, Ghent, Belgium
\and Ragn-Sells AB, Sweden
\and The Swedish e-Science Research Centre, Sweden \\
}

\maketitle 
\setcounter{footnote}{0}

\begin{abstract}
Increasing the circularity of resource use in our society has been recognized as a path to sustainability, i.e., transitioning into a more circular economy. There are many different circular strategies to do so, such as reusing products and components, refurbishing and remanufacturing used products, or recycling left-over or used materials. To enable these strategies, it is necessary to share information at the infrastructure level and to communicate between industry sectors along the product life cycle. Enabling semantic interoperability in this information sharing and communication is therefore a key to increasing circularity. However, knowledge representation for the circular economy (CE) domain, which involves many relevant industry sectors related to product life cycles, remains challenging. To bridge this gap, we developed the Circular Economy Ontology Network (CEON) within the Onto-DESIDE project. This ontology network aims to fill gaps in CE by defining cross-sectorial concepts and to enable semantics-aware data documentation. We demonstrate CEON through cross-industry data documentation scenarios spanning construction, electronics, and textile sectors.






\end{abstract}

\section{Introduction}
Circular Economy (CE) is, by the definition from the European Union,\footnote{\url{https://www.europarl.europa.eu/topics/en/article/20151201STO05603/circular-economy-definition-importance-and-benefits}, accessed 2026-04-01} a model of production and consumption. 
The goal of this model is to extend the life cycle of a product through sharing, leasing, reusing, repairing, refurbishing, and recycling existing materials and products as long as possible.
This transformation implies the need to accelerate the digital and green transition, shifting from a linear economy to a circular economy.
Enabling the necessary information flows across multiple actors in a circular economy context is a basic requirement for enabling circularity, since unknown materials and product contents are often \textit{showstoppers} when it comes to applying circular strategies. 
When the necessary information is shared transparently and with semantic accuracy among actors participating in product life cycles, more operational choices become available for handling products, e.g., processes implementing both reuse and recycling as well as other strategies. 
However, since the CE domain works across industry sectors and a product's life cycle involves a multitude of actors, it is challenging to enable the appropriate level of granularity for information to be shared among actors, with suitable semantics encoded. 

This paper presents the \textbf{\underline C}ircular \textbf{\underline E}conomy \textbf{\underline O}ntology \textbf{\underline N}etwork (CEON), an ontology network for cross-industry data documentation in the CE domain, which is one of the main outcomes from the Onto-DESIDE (Ontology-based Decentralized Sharing of Industry Data in the European Circular Economy) project.\footnote{Onto-DESIDE (Jun. 2022 -- Nov. 2025): \url{https://doi.org/10.3030/101058682}}
Specifically, we present the latest version (v1.0.0) of CEON, containing substantial updates\footnote{The changelog of CEON: \url{https://liusemweb.github.io/CEON/dev/changelog/}} in comparison with the initial version presented in~\cite{evaWOP2023}. 
The remainder of this paper is organized as follows. 
Section~\ref{sec-related-work} introduces the related ontologies and standardization based on a new survey extending~\cite{Li2023-CE-survey}.
Section~\ref{sec-ceon-method} introduces the methodology.
Section~\ref{sec-ceon} describes the details of CEON.
Section~\ref{sec-ceon-eval} presents the evaluation of CEON.
Finally, in Sections~\ref{sec-discussion} and~\ref{sec-conclusion}, we discuss the impact, reusability, and limitations of CEON, and present concluding remarks.


\section{Related Work}
\label{sec-related-work}
In the last five years, several EU-level research projects have investigated how to facilitate digitalization in industries with a sustainable economy purpose.
For example, the \textit{JIDEP} project\footnote{JIDEP (Jun. 2022 -- May 2025): \url{https://doi.org/10.3030/101058732}} focused on industrial data exchange and delivered a Material Passport Ontology~\cite{SEDDIQUI2026104465} focusing on interoperable data sharing and exchange at the material level.
The \textit{AUTO-TWIN} project\footnote{AUTO-TWIN (Dec. 2022 -- Nov. 2025): \url{https://doi.org/10.3030/101092021}} aimed at creating digital twins in circular value chains and reused the IDS (International Data Space) ontology~\cite{IDS-onto-2020} to enable data exchange between industries (e.g., manufacturing and product domains) within value networks.
The \textit{Circular TwAIn} project\footnote{Circular TwAIn (Jul. 2022 -- Jun. 2025): \url{https://doi.org/10.3030/101058585}} aimed to develop an AI platform supported by data sharing within manufacturing data spaces, and released the Circular TwAIn Ontology Library supporting technical components in the platform~\cite{Circular-TwAIn-2025}. 
The \textit{Reincarnate} project\footnote{Reincarnate (Jun. 2022 -- May 2026): \url{https://doi.org/10.3030/101056773}} focused on developing technical solutions for buildings, construction products and materials, and released the Reincarnate Ontology~\cite{samaneh_rezvani_2024_11354328}, which is built upon CEON and includes mappings to CEON.
Other existing work has developed ontologies for knowledge representation in the CE context. 
These ontologies include Circular Materials and Activities Ontology and Circular Exchange Ontology~\cite{sauter2019ceo}, Building Circularity Assessment Ontology~\cite{BCAO-2021}, and Sustainable Bioeconomy and Bioproducts Ontology~\cite{BiOnto-2021}, all of which have been investigated in our previous survey~\cite{Li2023-CE-survey}.
While these ontologies address CE knowledge representation, they are each confined to a single industry sector or technical platform and do not provide a unified, cross-industry framework.
Compared with them, CEON has a primary goal of enabling cross-industry data sharing, including a multi-granularity perspective spanning from resources and products to materials and their components, along with all relevant activities in circular value networks.
The \textit{diversity} among these recent and existing efforts aligns with the idea presented by Kirchherr et al.~\cite{ce-research-2023} that the concept of CE needs to be more specific and detailed in its operationalization.

Beyond ontology efforts, there are existing and ongoing standardization initiatives aimed at achieving \textit{uniformity} in CE, which are building standards at the global and EU levels. 
For instance, the recently established ISO (International Organization for Standardization) standard, \textit{ISO 59004:2024 Circular Economy – Terminology, Principles and Guidance for Implementation},\footnote{\label{iso-59004-2024}ISO 59004:2024: \url{https://www.iso.org/standard/80648.html}, accessed 2026-07-22} defines key terminology, proposes circular economy principles, and provides guidance for CE implementation.
The \textit{Ecodesign for Sustainable Products Regulation (ESPR)}\footnote{\label{espr-2024}ESPR: \url{https://eur-lex.europa.eu/eli/reg/2024/1781/oj/}, accessed 2026-07-22} came into force on July 18, 2024, aiming to provide a framework for defining ecodesign requirements for certain products.
Both have served as a reference and a non-ontological resource in CEON's development.
Moreover, the \textit{diversity} of ontology approaches and the effort toward \textit{uniformity} via standardization motivate the need for CEON as a network of ontologies.

\section{Methodology}
\label{sec-ceon-method}

The agile ontology engineering methodology, \textit{eXtreme Design (XD)}~\cite{presutti2009extreme,blomqvist2016engineering}, is combined with the \textit{Modular Ontology Modeling (MOMo)}~\cite{shimizu2023modular} to develop CEON.
The combination includes adopting a rapid prototyping and requirements-driven approach from XD, and the overall outline from MOMo such as identifying existing Ontology Design Patterns (ODPs)~\cite{blomqvist2005patterns,presutti2012pattern} and creating module diagrams.
The Onto-DESIDE project methodology and an analysis of the project requirements drive this combination.
For instance, the whole project had three iterations, each with an evaluation of the technical deliverables, including the ontology. 
Therefore, an agile methodology would fit in this case, which prioritizes flexibility, domain experts' requirements, and rapid, iterative delivery.
Moreover, the project requirements highlight the need for cross-domain modeling spanning topics such as resources, processes, circular value networks, and dedicated industry use cases, including construction, electronics, and textile.
An ontology network includes different modules, and general ODPs would address cross-domain interoperability to a great extent while maintaining suitable modeling granularity. 

We were aware of other ontology engineering methodologies such as \textit{LOT (Linked Open Terms)}~\cite{POVEDAVILLALON2022104755} and \textit{SAMOD}~\cite{SAMON2017}. 
However, XD is a more suitable one since it combines the ideas of both agile development and ODPs.
Although we did not explicitly follow LOT, which also has a focus on ontology publishing and documentation, we set up an ontology development and publishing pipeline\footnote{\label{ceon-git}CEON development/publishing repository: \url{https://github.com/LiUSemWeb/CEON}} to satisfy the FAIR (Findable, Accessible, Interoperable, and Reusable) principles~\cite{wilkinson2016fair}. Hence, the work is also generally in line with LOT.

\begin{table}[t!]
\centering
\footnotesize
\caption{Examples of circular enablers and corresponding CQs for developing CEON.}
\label{tab:ceon-req}
\renewcommand{\arraystretch}{1.1}
\setlength{\tabcolsep}{0.4pt}
\begin{tabular}{p{3.4cm}|p{8.7cm}}
\hline
\textbf{CE requirements}  & \textbf{Competency Question}                                               \\ \hline
The capacity to understand interrelations between processes and actors in the system.  & What are the connections and dependencies between actors and processes in a certain value network? \newline What are the energy components in this system?                 \\ 
\hline
The capacity to identify and consider all (relevant) system actors.                                                         & What are the actors (and their roles) in the value network? \newline What are the connections and dependencies between actors and materials/components/products in the value network?                                        \\
\hline
The capacity to consider processes throughout entire life cycle. &  What is the value network implementation in terms of circular strategies?  \newline What is the process breakdown of this life cycle? \\
\hline
The capacity to develop new configurations. & What are the characteristics, e.g., quality, of this material? \newline What kinds of value are involved in a collaboration/process? \\
\hline
\end{tabular}
\end{table}

\subsection{Requirements Analysis}
The Onto-DESIDE project requirements span across multiple perspectives including cross-domain challenges, circularity concepts, and bottom-up use-case requirements.
Taking a top-down perspective, 13 CE requirements (\textit{circular enablers}) were developed, based on general CE terminology and concepts used in standards and value flow analysis methodologies, such as \textit{Circularity Thinking}~\cite{blomsma2018circularity}. 
Conversely, taking a bottom-up perspective, 42 use-case requirements (13 for construction, 6 for electronics, and 23 for textile) were developed, and generalized to identify commonalities across industry domains. 
The requirements were then mapped to 9 topics, including circular value network, value, actor, process, material, product, location, quantities and units, and provenance. 
Based on this, we proposed relevant competency questions (CQ), contextual statements, and reasoning statements.
The latest set of ontological requirements includes 90 CQs across 13 CE requirements.
Additionally, we proposed use-case related CQs based on the use-case requirements for developing use-case ontologies.
Table~\ref{tab:ceon-req} shows example requirements and CQs.
More information about the methodology and the development of CEON is presented in the development documentation.\footnote{\label{ceon-dev-doc}CEON development documentation:  \url{https://liusemweb.github.io/CEON/dev/}}

\subsection{Ontology Development and Publishing}

We slightly adapted the XD methodology, i.e., creating a modified version of its design loop. 
One of the XD principles of ``pair design" was modified, since we did not have the resources (i.e., ontology engineers) to allow them to continuously work in pairs. 
Instead, we set up a method where ontology modules were created by one ontology engineer and then reviewed by another, in line with the idea of code reviews in software engineering. 
Thus, ontology engineers still work in pairs, but without the requirement of continuous synchronization of efforts. 

The development of the ontology network entailed multiple interdependent ontologies, most of which went through multiple development iterations. 
In order to keep track of such changes, we used a GitHub repository\footref{ceon-git} to handle versioning and to create new releases. 
The w3id service is used to provide permanent identifiers under a common namespace~(\url{http://w3id.org/CEON/}).

\section{Circular Economy Ontology Network}
\label{sec-ceon}

\begin{figure}[t!]

    \centering
    \includegraphics[width=1.0\textwidth]{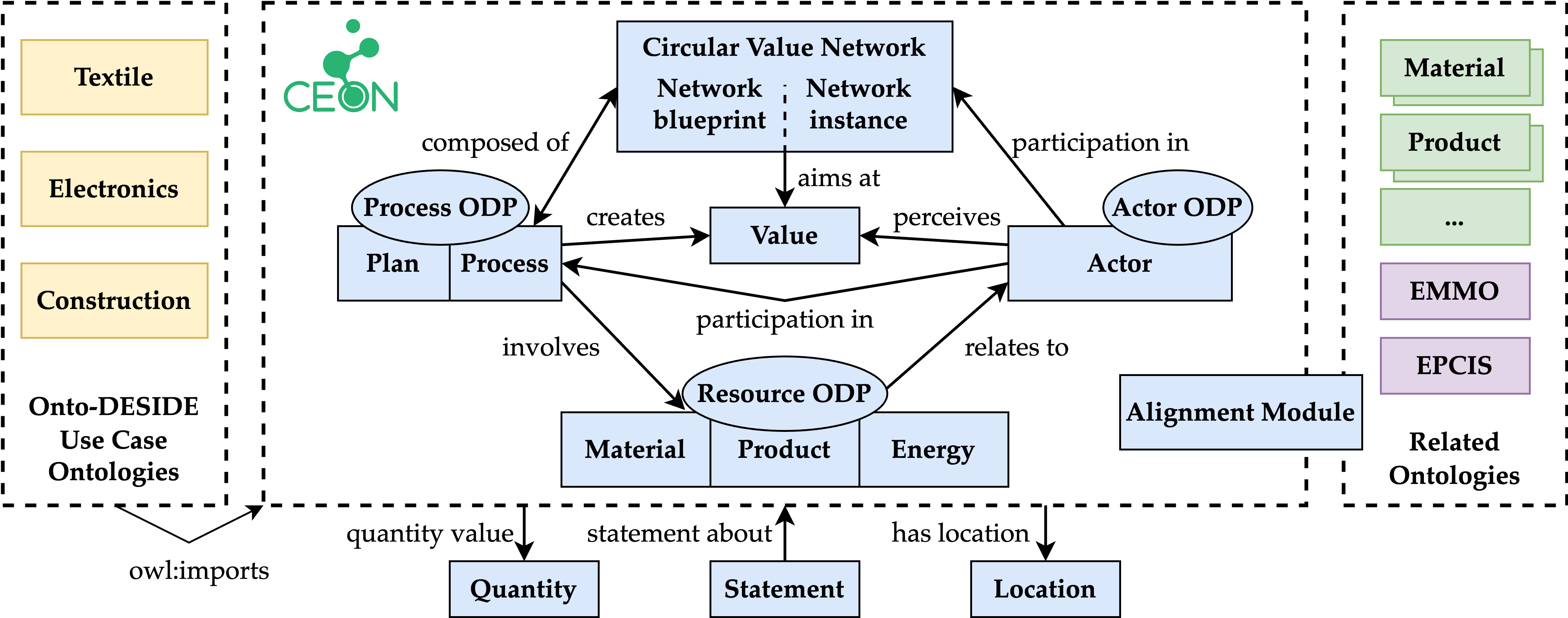}
    \caption{Informal illustration of the core topics of the ontology network.}
    \label{fig:ceon-network}
\end{figure}

The overview architecture of CEON is illustrated in Figure~\ref{fig:ceon-network}.
It includes the core modules for actor, circular value network, value, process, and resource, as well as general relationships among these modules.
Figures~\ref{fig:ceon-actor} to~\ref{fig:ceon-product} (based on the Chowlk visual notation~\cite{chowlk-2022}) show the basic classes and object properties in these modules.
Additionally, three modules for supplementary information are modeled, including quantity, statement, and location; three use-case ontologies were developed to reuse CEON's core modules. 
Furthermore, to enhance the semantic interoperability of CEON with other ontologies, we provide an alignment module that includes semantic mappings between CEON and related ontologies (more detailed alignment methodology and results are presented in~\cite{li-om-kgfs-2024} and~\cite{li-om-kgfs-2025}).

\subsection{Core Modules: Actor ODP and Actor}
\begin{figure}[t!]
    \centering
    \includegraphics[width=1.0\textwidth]{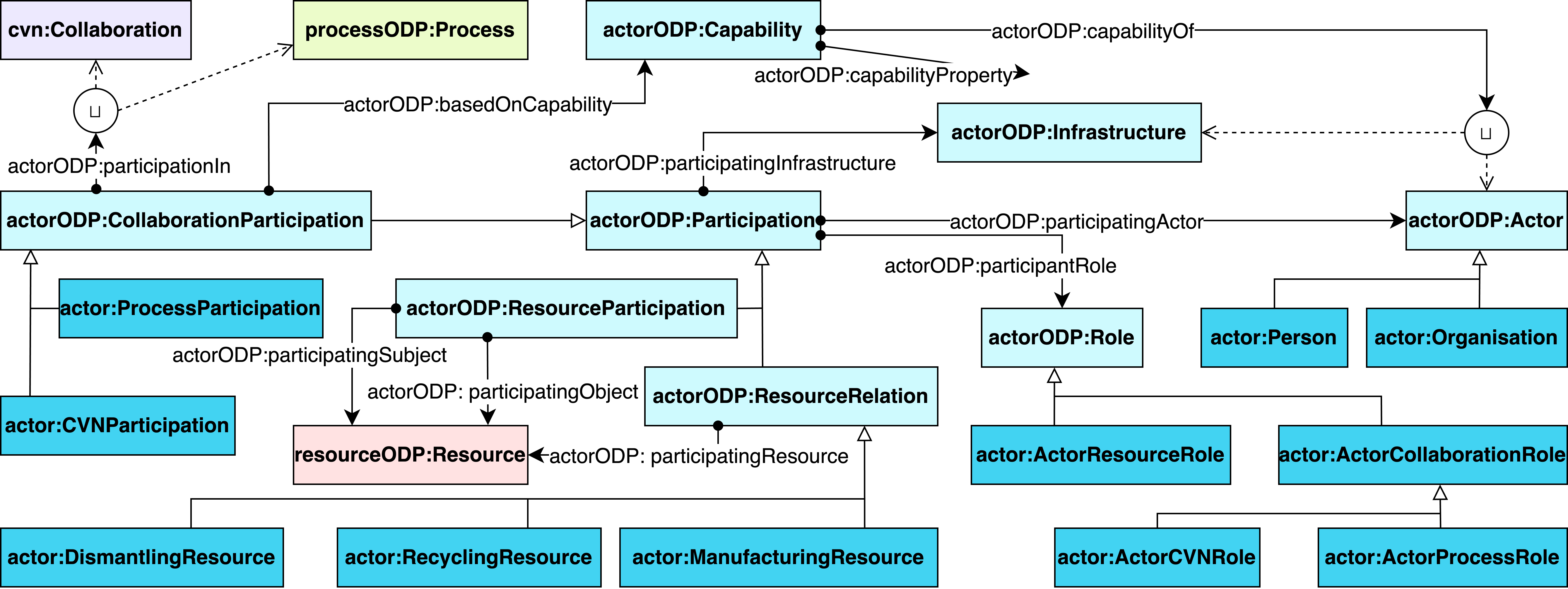}
    \caption{Actor ODP and Actor module.}
    \label{fig:ceon-actor}
\end{figure}

A circular value network is, in essence, composed of a set of actors filling certain roles in different phases of the network's flows and in relation to certain resources. 
Hence, the actors are the ones who actually realize the value network and perform the work to transform materials, products, and their components across the various phases of the value network. 
Actors can be modeled at two levels, i.e., as actor types that can fill certain typical roles in a network, such as a ``recycler" or ``manufacturer", and the concrete actors, which are usually organizations taking on those roles in a specific network instantiation (e.g., a recycling or manufacturing company). 
Actors are also related to their capabilities and competencies, which determine if they are able to fulfill a certain role in a network or not. 
Furthermore, actors take on various roles in relation to resources, e.g., holding, owning, selling, or buying them. 
Figure~\ref{fig:ceon-actor} shows the \textbf{Actor ODP} and its specialization, the \textbf{Actor} module.

The {Actor ODP} holds the most general concepts, essentially independent of any industry domain or circular strategy. 
It can be seen as a variant of the common participation\footnote{\url{https://odpa.github.io/patterns-repository/Participation/Participation}, accessed 2026-07-19} and participant role\footnote{\url{https://odpa.github.io/patterns-repository/ParticipantRole/ParticipantRole}, accessed 2026-07-19} ODPs. 
For instance, the core concept is \texttt{Participation}, which represents a participation linking an \texttt{Actor} to either a \texttt{Resource} or a \texttt{Process}, and the actor holds a specific  \texttt{Role}. These concepts are further specialized to represent more detailed participations such as  \texttt{ProcessParticipation} and \texttt{ResourceParticipation}.
Moreover, the Actor module includes specific roles related to circular strategies.
Such a way of modeling would naturally support many-to-many relationships between actors and resources or processes in a circular value network. 

\subsection{Core Modules: Circular Value Network and Value}
\begin{figure}[t!]

    \centering
    \includegraphics[width=1.0\textwidth]{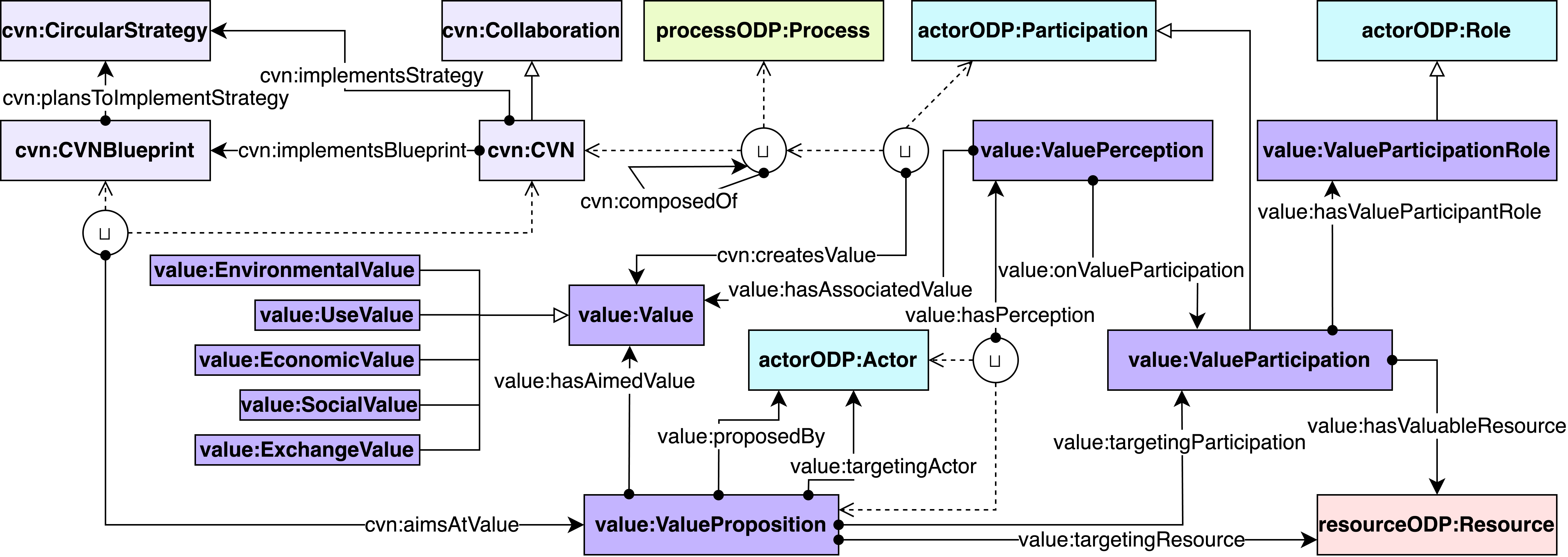}
    \caption{CVN and Value modules.}
    \label{fig:ceon-cvn-value}
\end{figure}

The \textbf{CVN} and \textbf{Value} modules, as shown in Figure~\ref{fig:ceon-cvn-value}, detail the core concepts about \texttt{Circular Value Network (CVN)} and \texttt{Value}. 
Our starting point for the {CVN} module was an analysis of relevant terminology, ontologies, and emerging standards (i.e., ISO 59004:2024\footref{iso-59004-2024} and ESPR\footref{espr-2024}) as well as a generalization over the use cases and project requirements in Onto-DESIDE.
The CVN module reuses and connects to concepts defined in the actor, process, and value modules, as a \texttt{CVN} inherently links actors through processes to the creation of value.
A \texttt{CVN} operates according to a \texttt{CVNBlueprint}, which describes the planned setup of the network.  
A planned setup usually specifies how resources are transformed or operated by actors of certain types, follows or targets certain types of circular strategies (e.g.,  refurbishment of a product), and relates to typical value propositions and goals. 
We separate \texttt{CVNBlueprint} from the concrete \texttt{CVN} instance because circular value networks are first designed as templates and then instantiated with specific actors, and mixing the two would complicate data documentation.
The {CVN} module supports modeling concrete instances of value networks, i.e., actual value networks where the roles are filled by various actors of the appropriate types (through \texttt{Participation} as shown in Figure~\ref{fig:ceon-actor}), with a specific goal and value proposition in mind. 

Although value is a central concept in the CE domain and closely related to the circular value network through its value proposition, it remains challenging to define and formally represent in data. 
Following discussions on the value concept currently in progress in other forums, e.g., standardization bodies, the \textbf{Value} module models the basic concepts and relationships to represent values, value propositions, value perceptions, and related constructs.
In addition, we model the connections between the Value module and the CVN module by defining that specific CVNs or their blueprints can aim at specific values, thereby representing value propositions. 
Moreover, we capture the fact that value is contextual, e.g., a resource has value in relation to a specific actor or in a specific context (with targeting actors, resources, or participations). 

\subsection{Core Modules: Process ODP, Process and Plan}

\begin{figure}[t!]
    \centering
    \includegraphics[width=1.0\textwidth]{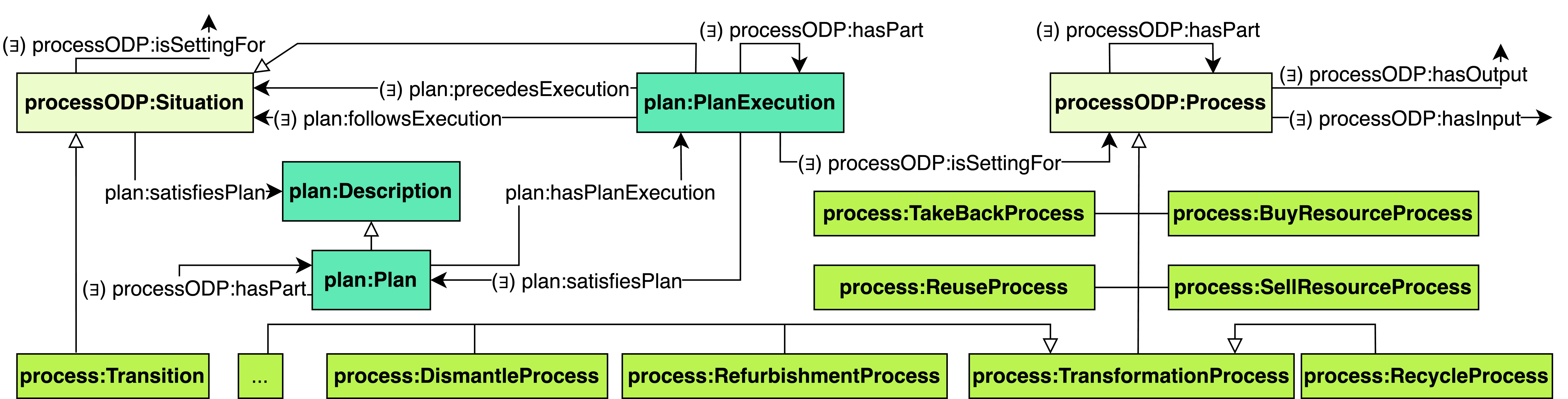}
    \caption{Process ODP, Process and Plan Modules.}
    \label{fig:ceon-process}
\end{figure}

Figure~\ref{fig:ceon-process} illustrates the concepts and relationships about processes in CEON.
The realization of this topic consists of the \textbf{Process ODP}, specifying the generic concepts involved in process modeling, and the \textbf{Process} module as a specialization of that, to include the CVN-specific processes that are targeted in the project, and the \textbf{Plan} module detailing process settings and plan executions. 
The modeling is inspired by the TaskExecution ODP\footnote{\url{https://odpa.github.io/patterns-repository/TaskExecution/TaskExecution}, accessed 2026-07-19} to distinguish a plan and its execution, and the Transition ODP\footnote{\url{https://odpa.github.io/patterns-repository/Transition/Transition},\\ accessed 2026-07-19} to represent a process taking some input and transforming it to some output, i.e., moving the current state from one situation (context) to another.
Each circular value network realizes one or more circular value flows, which can be seen as a process of transforming certain resources, e.g., from materials into products, and then potentially back again. 
Such processes have different phases, such as transforming materials to components or deconstructing a product into its material composition. 
Furthermore, each phase can be divided into smaller steps (i.e., pieces of work), which different actors can perform. 
We model all of these at the execution and sub-execution levels, specializing the general process concept into one that can transform one situation into another, for example, by changing the state of affairs, such as the situations of actors or resources.
Then each situation is designed to satisfy a plan, which has a corresponding plan execution. 
Each step may then also have inputs and outputs of resources, with respect to the situation of its corresponding process.
This modeling reflects that CVN processes often involve state changes that cannot be fully captured by listing inputs and outputs alone.

\subsection{Core Modules: Resource ODP, Material, Product and Energy}

\begin{figure}[t!]
    \centering
    \includegraphics[width=1.0\textwidth]{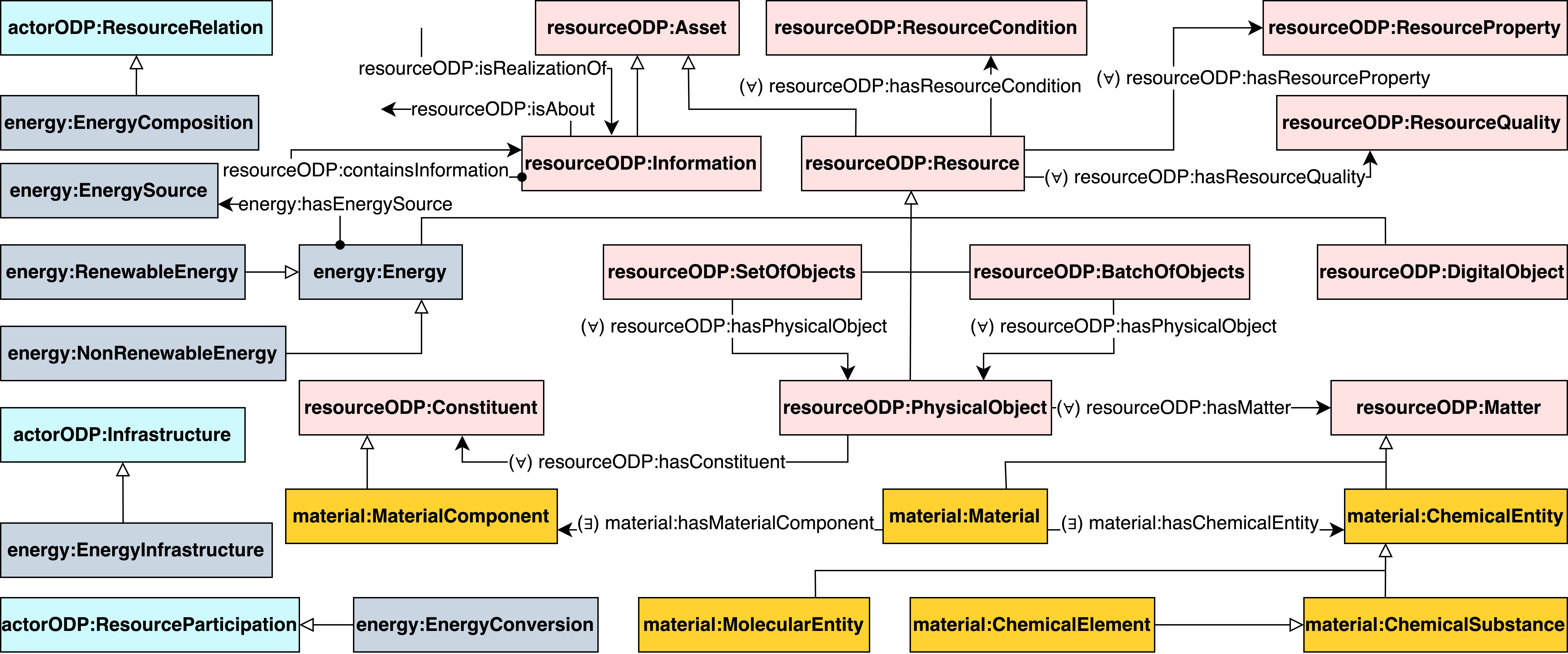}
    \caption{Resource ODP, Material and Energy Modules.}
    \label{fig:ceon-resource}
\end{figure}

\begin{figure}[ht!]
    \centering
    \includegraphics[width=1.0\textwidth]{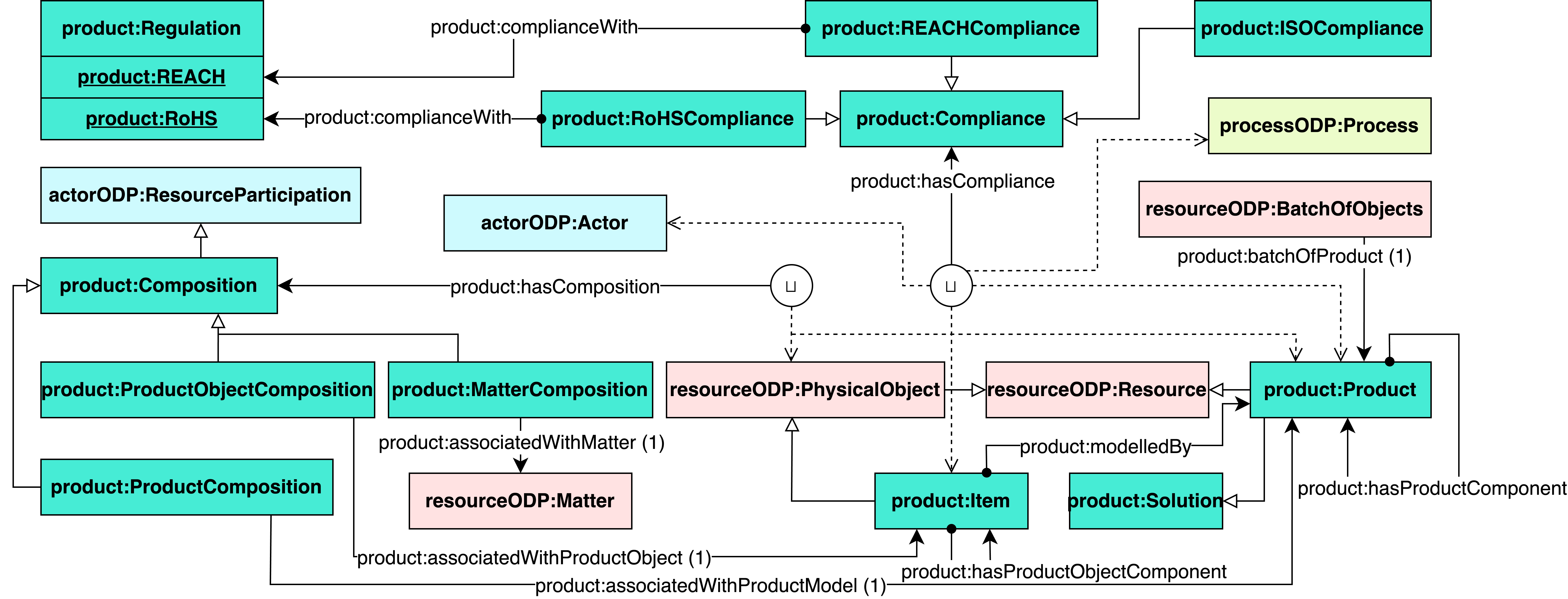}
    \caption{Product Module.}
    \label{fig:ceon-product}
\end{figure}

Resources are at the core of the circular economy, since they constitute the physical flows that are circulated, and the things that are needed as input and output of each process. 
Most prominently, the resources are the materials, products, and their components that the network aims to manage in a circular manner. 
Additionally, resources can include the materials needed for processing, such as consumables or catalysts, as well as energy required for different processes.

This part of the network is realized through a generic \textbf{Resource ODP}, which is then specialized into three modules: \textbf{Material}, \textbf{Energy}, and \textbf{Product}.
Figure~\ref{fig:ceon-resource} shows the concepts and relationships in the Resource ODP, Material, and Energy modules, while Figure~\ref{fig:ceon-product} illustrates the product module. 
\texttt{Resource} has a detailed breakdown including sub-classes such as \texttt{PhysicalObject}, \texttt{Matter}, and \texttt{Constituent}, which are further extended in the Material and Product modules.
\texttt{Resource} can have three specific relations to represent a resource's condition, property, and quality. 
These are generalized from the three use cases in the project, which reflect common modeling concerns in CE.
Moreover, the Material module is modeled in the same style as the \textit{EMMO (Elementary Multiperspective Material Ontology)}~\cite{emmo-2024} for materials modeling regarding \texttt{ChemicalEntity} and its detailed sub-classes. 
This is because EMMO provides a well-established 
and scientifically grounded representation of chemical entities that avoids  duplication and promotes interoperability with existing materials science ontologies.
The Energy module is modeled at a general level, including basic concepts, \texttt{Energy}, \texttt{EnergySource}, \texttt{EnergyComposition}, and \texttt{EnergyConversion} to represent the role of energy in circular value networks via participations.
The Product module is modeled at two levels: the abstract level, a product in terms of its model, and the physical level, a specific product item. 
For instance, \textit{LEGO Star Wars - Millennium Falcon} represents a product model (\texttt{Product}), while a Star Wars fan may purchase a specific product (\texttt{Item}) of that model.
Furthermore, the Product module models several concepts and relationships to represent compliance and its corresponding regulations of actors, processes, or products.

\subsection{Supplementary Modules and Use-Case Ontologies}
\paragraph{\textbf{Quantity Module.}}
A cross-cutting notion for sharing data in the circular economy domain is to represent actual quantities, such as material or product quantities, energy use, or costs. 
To represent more detailed quantity information for resources or processes, we modeled the {Quantity} module to represent quantity values, for instance, those associated with processes such as dismantling or transport costs. 
We reused the \textit{QUDT (Quantities, Units, Dimensions, and Types)} ontology\footnote{QUDT ontology: \url{https://qudt.org}, accessed 2026-05-04}, e.g., by specializing the \texttt{quantityValue} object property.

\paragraph{\textbf{Location Module.}}
Location appears in many places in the overall list of requirements. 
Resources are associated with a specific location at a given point in time. 
However, they may also have an origin point and a traceable history of places they have been, similarly for actors, information, etc. 
Different use cases demand varying levels of granularity in location information.
For example, a construction-related use case might require precise details, such as specifying that an object is on the second floor of a building.
In contrast, a take-back system may require precise coordinates for a crate of products awaiting pickup.
Therefore, we modeled a dedicated module for the relevant concepts and relationships of location. 
This module integrates existing standard ontologies (e.g.,  the standardized \textit{GeoSPARQL}~\cite{battle2012enabling} from the Open Geospatial Consortium) and introduces new concepts and relationships relevant to CEON. 

\paragraph{\textbf{Statement Module.}}
From the requirements of Onto-DESIDE, it is also clear that we need to be able to express facts about facts, i.e., metadata about the information shared in the circular value network. 
The most obvious case is to keep the traceability of what actor has made a certain claim, e.g., about a product or its components, or any resource in general. 
Thus, we model the general concept \texttt{Statement} and the relationship \texttt{statementAbout} in the Statement module.
This module is further enhanced to represent statements following the \textit{PCDS (Product Circularity Data Sheet)}, which is a recently established ISO standard\footnote{ISO 59040:2025:  \url{https://www.iso.org/standard/82339.html}, accessed 2026-07-22} (\textit{ISO 59040:2025 Circular economy — Product circularity data sheet}).

\paragraph{\textbf{Use-Case Ontologies.}}
In addition to the core and supplementary modules, we developed three use-case ontologies in the domains of construction, electronics, and textile with respect to the industry partners of Onto-DESIDE. 
The development of these use-case ontologies, each of which has domain requirements, followed the same methodology as introduced in Section~\ref{sec-ceon-method}.
More details are presented in the development documentation.\footref{ceon-dev-doc}

\section{Evaluation of CEON}
\label{sec-ceon-eval}
The evaluation of CEON is conducted in each iteration of Onto-DESIDE, focusing on both establishing the intrinsic ontology quality and the extrinsic ontology usage.
For the former, the evaluation dimensions cover ontology characteristics, consistency, adherence to modeling best practices, requirement fulfillment, and coverage (by verifying CQs, see Section~\ref{sec-technical-eval}).
For the latter, the dimensions cover cross-domain data documentation for the CE domain (see Section~\ref{sec-cross-application}) and semantic interoperability among CE-related ontologies.

\subsection{Technical Evaluation}
\label{sec-technical-eval}
\paragraph{\textbf{Requirements Coverage.}}
For evaluating CEON in terms of requirements fulfillment and coverage, we followed a testing method similar to~\cite{blomqvist2012ontology}, where CQs are verified by formulating SPARQL queries. 
However, since Onto-DESIDE did not require or collect industry data throughout the ontology development process, populating the ontology network with real instances was not feasible.
Therefore, we have settled for testing whether it is possible to formulate a SPARQL query that corresponds to the CQ, rather than also adding actual test data and running the query. 
The reason is that to do the latter, we would also have to create “example data” (synthetic data modeled according to the ontology) ourselves, which would introduce a large bias, and we would likely not find many more mistakes by doing so.
The final evaluation of requirement coverage showed 25 CQs fully covered, 28 partly covered, and 37 uncovered out of 90 CQs.
Partial coverage indicates that only some aspects are currently modeled, and some parts are still missing or not intended for inclusion in the core modules, but are specific to an industry domain. 
The evaluation results show that 41\% of CQs are not covered, which we attribute to three categories of reasons. First, certain requirements remain at a strategic or goal level of the CE domain that does not correspond to representable knowledge (e.g., \textit{What is the best option for energy replacement?}), and are therefore intentionally outside the scope of the core modules. Second, some CQs are not covered in the core modules because they are specific to a single industry domain and are addressed instead in the corresponding use-case ontologies. Third, a small number of CQs reflect modeling areas that are still under active standardization or research, such as detailed value representation, and will be addressed in future versions as standards and research mature.

\paragraph{\textbf{Pitfall Checking.}}
We employed the OntOlogy Pitfall Scanner (OOPS!)~\cite{poveda2014oops} and OOPS! for FAIR (FOOPS!)~\cite{foops2021} to identify potential issues requiring attention. 
OOPS! identifies issues at three levels including critical, important, and minor~\cite{poveda2014oops}.
For instance, a cyclic definition in the concept hierarchy will raise a critical pitfall.
Missing domain or range definitions of properties are important issues while missing annotations are minor ones.
In the evaluation of CEON using OOPS!, our strategy is to address all critical issues identified and leave important or minor issues for further discussion. 
This is because we intentionally keep modeling flexibility, such as leaving the domain or range of certain properties undefined.
Furthermore, we used FOOPS! to check if any important metadata is missing, because we want to ensure that CEON contains sufficient metadata and is published according to the FAIR principles~\cite{wilkinson2016fair}.
For instance, this includes using a persistent and resolvable URI (Findable), providing different serialization formats (Accessible), using existing vocabularies for metadata annotations (Interoperable), and providing human-readable documentation (Reusable).

\subsection{Cross Domain Data Documentation}
\label{sec-cross-application}

\paragraph{\textbf{Cross-Industry Setup.}}
The cross-industry setup is based on a tangible example of the construction of a space (e.g., a room) and involves elements from different sectors.
For instance, construction elements such as doors and tiles, textiles used for acoustic insulation, and electronics like sound systems for alarms or other uses are all considered. 
The scenario covers three life stages of the constructed space and relevant circular strategies as discussed in~\cite{blomsma2018circularity}.

\begin{tcolorbox}[colback=framegray,colframe=lightblue, title= Scenario Overview,
fonttitle=\footnotesize\bfseries,
boxsep=1.5pt, top=0.5mm, bottom=0.5mm, left=2mm, right=2mm]
  \textbf{\textsl{Beginning-of-Life (Recycle):}} A building owner wants to buy floor tiles from a construction company, which therefore needs to buy material for manufacturing acoustic layers of tiles. Two actors could offer such material: a recycling company that recycles shoes and sells the soles and laces as feedstock, and a specific material supplier selling virgin material.
  
  \textbf{\textsl{Middle-of-Life (Maintenance):}} The building owner sees that the speaker of a sound system is broken. The owner accesses the repair instructions and discovers that the original equipment manufacturer offers a repair service. Therefore, the building owner sends  the speaker back to the manufacturer and receives an updated version.

  \textbf{\textsl{End-of-Life (Reuse and Take Back):}} The building owner decides to tear down a building. The doors from the building are offered for sale. There are different circular strategies for the owner. For instance, the doors can be dismantled or recycled; they can be sold to another actor or taken back by the manufacturer.    
\end{tcolorbox}

\begin{figure}[t!]

    \centering
    \includegraphics[width=1.0\textwidth]{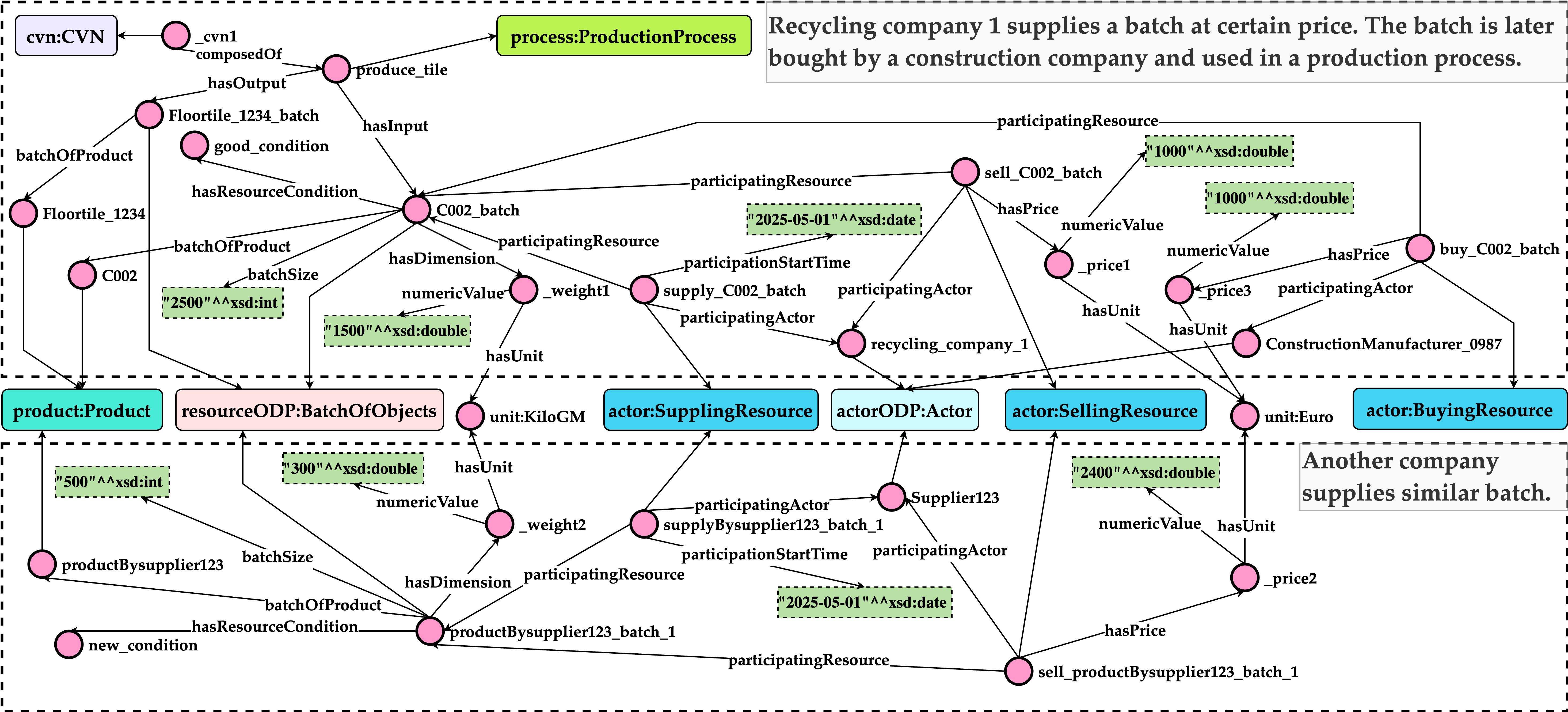}
    \caption{A recycling example of a batch of products (beginning-of-life).}
    \label{fig:1_recycling}
\end{figure}

\begin{figure}[t!]

    \centering
    \includegraphics[width=1.0\textwidth]{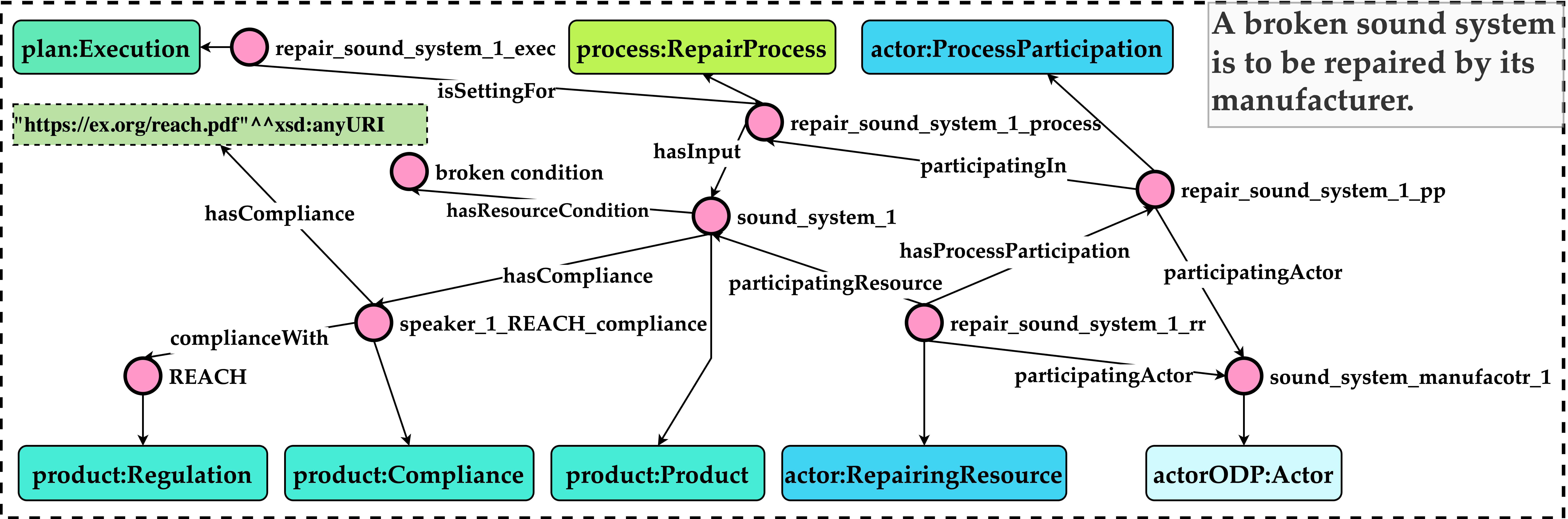}
    \caption{A repairing example of a set of products (middle-of-life).}
    \label{fig:2_repairing}
\end{figure}

\begin{figure}[t!]

    \centering
    \includegraphics[width=1.0\textwidth]{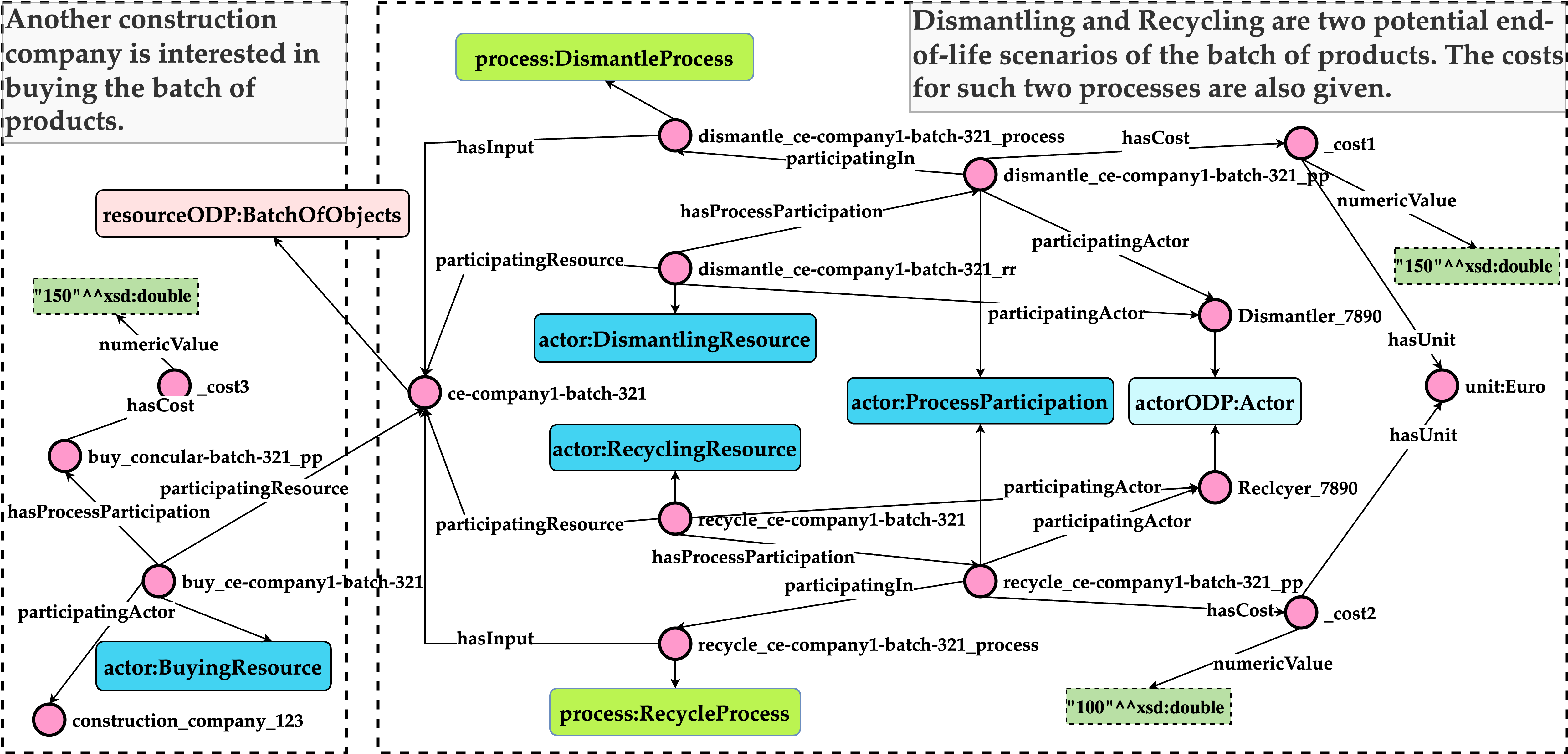}
    \caption{An example of handling a batch of products (end-of-life).}
    \label{fig:3_reusing}
\end{figure}

These examples depict scenarios across a product's life cycle, including purchasing products from a recycling company, repairing products by manufacturers, and handling products at the end of their life (e.g.,  through dismantling, recycling, or take-back programs).
Sharing information transparently with semantics encoded is important to enable the circular strategies. 
Figure~\ref{fig:1_recycling} shows an instantiation where a buyer compares prices of similar products offered by two different suppliers, purchasing one from the recycling company. These recycled products are then used in a production process to manufacture new goods, representing a beginning-of-life scenario.
Figure~\ref{fig:2_repairing} illustrates an instantiation where a product is found to be broken. 
In such cases, the original manufacturer may offer a repair service.
Figure~\ref{fig:3_reusing} presents an end-of-life scenario where a product may be dismantled, recycled, or sold to other actors.
The resources in these scenarios, such as floor tiles, speakers, and doors, can also be typed using concepts from the corresponding use-case ontologies.

\paragraph{\textbf{SPARQL Query Examples.}}
We show three SPARQL query examples corresponding to the cross-industry scenarios. 
For the beginning-of-life scenario, Listing~\ref{lst:sparql-beginning} queries which product batches are available for purchase and at what price and conditions, providing the information a construction company needs to compare recycled versus virgin material suppliers. 
For the middle-of-life scenario, Listing~\ref{lst:sparql-middle} queries the compliance information of a product being sent for repair, including the company, resource, applicable regulation, and certificate, which is essential for verifying that a repair service meets the required standards. 
For the end-of-life scenario, Listing~\ref{lst:sparql-end} queries the potential circular operations available for a product and their associated costs, including the type of operation (e.g., dismantling, recycling, resale), cost, resource, and responsible actor, which supports the owner's decision on how to handle end-of-life products.
These queries can be tested directly using the public SPARQL endpoint available at \url{https://liusemweb.github.io/CEON/sparql/}.

\begin{lstlisting}[float=h!, caption={SPARQL query based on the recycling example as shown in Figure~\ref{fig:1_recycling}.}, label={lst:sparql-beginning}]
PREFIX qudt: <http://qudt.org/schema/qudt/>
PREFIX actorODP: <http://w3id.org/CEON/ontology/actorODP/>
PREFIX resourceODP: <http://w3id.org/CEON/ontology/resourceODP/>
PREFIX product: <http://w3id.org/CEON/ontology/product/>
PREFIX quantity: <http://w3id.org/CEON/ontology/quantity/>
PREFIX actor: <http://w3id.org/CEON/ontology/actor/>

SELECT ?s ?resource ?product ?condition ?price ?unit WHERE {
  ?s a actor:SellingResource .
  ?s actorODP:participatingActor ?company .
  ?s actorODP:participatingResource ?resource .
  ?s quantity:hasPrice [
    qudt:numericValue ?price ; 
    qudt:hasUnit ?unit
  ] .
  ?resource product:batchOfProduct ?product .
  ?resource resourceODP:hasResourceCondition ?condition .
}                                                         
\end{lstlisting}

\begin{lstlisting}[float=h!, caption={SPARQL query based on the repairing example as shown in Figure~\ref{fig:2_repairing}.}, label={lst:sparql-middle}]
PREFIX actorODP: <http://w3id.org/CEON/ontology/actorODP/>
PREFIX product: <http://w3id.org/CEON/ontology/product/>
PREFIX actor: <http://w3id.org/CEON/ontology/actor/>

SELECT ?s ?company ?resource ?regulation ?certificate WHERE {
  ?s a actor:RepairingResource .
  ?s actorODP:participatingActor ?company .
  ?s actorODP:participatingResource ?resource .
  ?resource product:hasCompliance [
      product:complianceWith ?regulation ; 
      product:hasCertificate ?certificate
  ] .
}                                                        
\end{lstlisting}

\begin{lstlisting}[float=h!, caption={SPARQL query based on the end-of-life scenario as shown in  Figure~\ref{fig:3_reusing}.}, label={lst:sparql-end}]
PREFIX actorODP: <http://w3id.org/CEON/ontology/actorODP/>
PREFIX actor: <http://w3id.org/CEON/ontology/actor/>
PREFIX quantity: <http://w3id.org/CEON/ontology/quantity/> 
PREFIX qudt: <http://qudt.org/schema/qudt/> 

SELECT ?potentialDecision ?cost ?unit ?resource ?company WHERE {
  ?participation a actor:ProcessParticipation .
  ?participation quantity:hasCost [ 
    qudt:numericValue ?cost ; 
    qudt:hasUnit ?unit 
  ] .
  ?resourcerelation actor:hasProcessParticipation ?participation .
  ?resourcerelation actorODP:participatingResource ?resource .
  ?resourcerelation actorODP:participatingActor ?company .
  ?resourcerelation a ?potentialDecision . 
}
\end{lstlisting}

\section{Discussion}
\label{sec-discussion}
\paragraph{\textbf{Impact.}} 
To our knowledge, CEON is the first ontology network with a primary focus on cross-industry knowledge representation for data documentation in the CE domain.
Together with other technical contributions, such as the \textit{Open Circularity Platform}\footnote{Solid~\cite{solid2021} based platform: \url{https://onto-deside.ilabt.imec.be}}~\cite{DeMulder2024OpenCircularity}, Onto-DESIDE showcases decentralized data sharing for industries in the CE domain.
To further improve semantic interoperability, we provide semantic alignments between CEON and other CE-related and Digital Product Passport (DPP)-related ontologies, and we set up a new CE track at the Ontology Alignment Evaluation Initiative (OAEI) to evaluate ontology matching tools~\cite{li2025cetrack,oaei-2025-paper}.
Other work reuses CEON for applications such as knowledge graph construction~\cite{GALLINA2025643}, investigation of semantic interoperability for the CE and DPP domains~\cite{samaneh_rezvani_2024_11354328,bernier_2026_19885638}, and knowledge representation in the flat glass industry~\cite{EB-kg4s2026-glass}.

\paragraph{\textbf{Reusability.}} 
In addition to the public ontology documentation generated using pyLODE,\footnote{Ontology documentation tool: \url{https://github.com/rdflib/pyLODE}} the Onto-DESIDE project provides training materials\footnote{Onto-DESIDE training guides: \url{https://ontodeside.eu/training/}} that include ontology training, which will further improve the reusability of CEON. 
They can provide a starting point for users without experience in ontology modeling and use. 
Moreover, we provide the development documentation\footref{ceon-dev-doc} of CEON (including a maintenance plan), which will guide future users.

\paragraph{\textbf{Limitations and Future Work.}}
As mentioned in Section~\ref{sec-technical-eval}, some CQs from the user stories that are drawn out by the project requirements are not captured yet. 
This is because these ``Competencies" remain at the strategic level or are goals of the CE domain rather than knowledge representation. 
This aligns with the current situation, where the CE domain is still under development and new requirements may arise and evolve. 
In the future, we will integrate CEON in the different perspectives of CE applications, such as studying the needs
and solutions for interacting with, understanding, and using CE ontologies for domain experts, given that technical artifacts (e.g., ontologies) may not be directly usable or understandable to them.

\section{Concluding Remarks}
\label{sec-conclusion}
This paper presents the latest version of the Circular Economy Ontology Network (CEON), which aims to enable cross-industry knowledge representation and data documentation in circular value networks. The iterative development of CEON aligns with well-known ontology engineering methodologies and practices, and follows the Onto-DESIDE project's iterations and evaluations. 
In this paper, we showcase cross-industry scenarios where CEON is used for data documentation. 
As the CE domain continues to evolve with new industry requirements or standards, CEON is designed to grow alongside it, providing a semantic foundation for emerging cross-industry data sharing and knowledge representation needs.

\paragraph{\textbf{Acknowledgements.}}
This work has been financially supported by the EU Horizon project Onto-DESIDE (Grant Agreement 101058682), the Vinnova-funded projects Trace4Value (Dnr. 2021-04323) and SwePass (Dnr. 2024-02504), as well as the strategic research area Security Link. 
We would like to extend our gratitude to all partners in the Onto-DESIDE consortium for discussions throughout the ontology development, and to Professor Oscar Corcho for providing comments and feedback on the ontology network.

\paragraph*{\textbf{Resource availability statement:}} Source code is available at \url{https://github.com/LiUSemWeb/CEON}. The permanent IRI of the ontology is \url{https://w3id.org/CEON}. A SPARQL endpoint for testing example queries is available at \url{https://liusemweb.github.io/CEON/sparql/}. Furthermore, the development documentation is available at \url{https://liusemweb.github.io/CEON/dev/}.

\paragraph{\textbf{Declaration of use of Generative AI:}}
The authors used Anthropic Claude (Sonnet 4.6 and 5) to perform grammar and spell checking. The authors reviewed and edited the content as needed to take full responsibility for its content.
%
%
%
\bibliographystyle{splncs04}
\bibliography{ref.bib}
\end{document}